\documentclass[11pt]{article}

\usepackage[final]{acl}

\usepackage{times}
\usepackage{latexsym}

\usepackage[T1]{fontenc}

\usepackage[utf8]{inputenc}

\usepackage{microtype}

\usepackage{inconsolata}

\usepackage{graphicx}

\usepackage{hyperref}
\usepackage{url}
\usepackage{tabularx}
\usepackage{graphicx}
\usepackage{enumitem}
\usepackage{algorithm}
\usepackage{algorithmic}
\usepackage{amsmath}  
\usepackage{amsthm}
\usepackage{multirow}
\usepackage{bm}
\usepackage{mathrsfs}
\usepackage{placeins}
\usepackage{caption}

\usepackage{subfigure}

\usepackage[utf8]{inputenc} 
\usepackage[T1]{fontenc}    
\usepackage{hyperref}       
\usepackage{booktabs}       
\usepackage{amsfonts}       
\usepackage{nicefrac}       
\usepackage{microtype}      
\usepackage{xcolor}         

\usepackage{tabularray}
\usepackage{array}

\usepackage{multirow}
\usepackage[table,dvipsnames]{xcolor}

\usepackage{arydshln}
\usepackage{colortbl}
\definecolor{lightpurple}{rgb}{0.941, 0.941, 1.000} 
\usepackage[most]{tcolorbox}  

\newtcolorbox[auto counter]{promptbox}[2][]{
  enhanced,                 
  colback=lightpurple,
  colframe=gray,
  fontupper=\normalsize,         
  fonttitle=\bfseries\normalsize,
  title=Prompt~\thetcbcounter: #2,
  #1
}

\definecolor{darkblue}{rgb}{0, 0, 0.5}
\hypersetup{colorlinks=true, citecolor=darkblue, linkcolor=darkblue, urlcolor=darkblue}

\newcommand{\think}[1]{\textcolor{blue}{\texttt{<think>}} #1 \textcolor{blue}{\texttt{</think>}}}
\newcommand{\call}[1]{\textcolor{cyan}{\texttt{<tool\_call>}} #1 \textcolor{cyan}{\texttt{</tool\_call>}}}
\newcommand{\response}[1]{\textcolor{brown}{\texttt{<tool\_response>}} #1 \textcolor{brown}{\texttt{</tool\_response>}}}
\newcommand{\answer}[1]{\textcolor{purple}{\texttt{<answer>}} #1 \textcolor{purple}{\texttt{</answer>}}}

\newcommand{\gray}[1]{{\color{gray}{#1}}}

\title{TSPO: Breaking the Double Homogenization Dilemma in Multi-turn Search Policy Optimization}

\author{
\textbf{Shichao Ma}$^{1,2}$\thanks{Work done during internship at Ant Group.} \quad
\textbf{Zhiyuan Ma}$^{2}$ \quad
\textbf{Ming Yang}$^{3,2*}$ \quad
\textbf{Xiaofan Li}$^{4,2*}$ \quad
\textbf{Xing Wu}$^{2}$ \\
\textbf{Jintao Du}$^{2}$ \quad
\textbf{Yu Cheng}$^{2}$ \quad
\textbf{Weiqiang Wang}$^{2}$ \quad
\textbf{Qiliang Liu}$^{5}$ \quad
\textbf{Zhengyang Zhou}$^{1}$\thanks{Corresponding author.}
\textbf{Yang Wang}$^{1\dagger}$\\
\\
$^1$University of Science and Technology of China\quad
$^2$Tiansuan Lab, Ant Group Co., Ltd.\quad
$^3$Fudan University\quad \\
$^4$East China Normal University\quad
$^5$Central South University\\
}

\begin{document}
\maketitle
\begin{abstract}
Multi-turn tool-integrated reasoning enables Large Language Models (LLMs) to solve complex tasks through iterative information retrieval. However, current reinforcement learning (RL) frameworks for search-augmented reasoning predominantly rely on sparse outcome-level rewards, leading to a "Double Homogenization Dilemma." This manifests as (1) \textit{Process homogenization}, where the thinking, reasoning, and tooling involved in generation are ignored. (2) \textit{Intra-group homogenization}, coarse-grained outcome rewards often lead to inefficiencies in intra-group advantage estimation with methods like Group Relative Policy Optimization (GRPO) during sampling.
To address this, we propose \textbf{T}urn-level \textbf{S}tage-aware \textbf{P}olicy \textbf{O}ptimization (\textbf{TSPO}). 
TSPO introduces the First-Occurrence Latent Reward (FOLR) mechanism, allocating partial rewards to the step where the ground-truth answer first appears, thereby preserving process-level signals and increasing reward variance within groups without requiring external reward models or any annotations. Extensive experiments demonstrate that TSPO significantly outperforms state-of-the-art baselines, achieving average performance gains of \textbf{24\%} and \textbf{13.6\%} on Qwen2.5-3B and 7B models, respectively. 
Code is available at \
{
\color{lightpurple}
\url{https://github.com/Flipped-May/TSPO}}
.

\end{abstract}

\section{Introduction}

Large Language Models (LLMs) have recently shown strong abilities in complex reasoning through multi-turn interactions with external tools such as search engines, calculators, and code interpreters~\cite{team2025tongyi, jin2025search, chai2025parl}.
Unlike single-step decision-making, this multi-turn paradigm allows models to decompose complex tasks into a series of manageable sub-tasks. By iteratively retrieving and integrating information, LLMs have achieved remarkable success in challenging tasks like open-domain question answering (QA) and mathematical reasoning~\cite{zhang2025criticsearch, design2025reinforcing}.


Despite their potential, current reinforcement learning (RL) frameworks for multi-turn tool calling predominantly rely on outcome-level reward signals~\cite{jin2025search, chen2025learning}. In tasks like QA, performance is typically measured by Exact Match (EM)~\cite{maalouly2022exact}, where a binary reward is assigned only at the final turn. While this avoids potential \textit{reward hacking}~\cite{amodei2016concrete}, \textbf{\textit{it compresses the entire multi-turn dynamic tooling and reasoning process into a single scalar}}, obfuscating the quality of intermediate steps. We identify that this reliance on sparse, outcome rewards creates a systemic bottleneck, which we term the \textbf{Double Homogenization Dilemma}.

The first facet is \textbf{process-level reward homogenization}, where trajectories with diverse intermediate reasoning or retrieval quality receive identical outcome rewards. This phenomenon ignores progress made before the final step and incorrectly penalizes beneficial turn-level actions, such as successful information acquisition. The second facet is \textbf{intra-group reward homogenization}, which is particularly problematic in Group Relative Policy Optimization (GRPO) \cite{shao2024deepseekmath}. Since the binary nature of rewards often results in uniform-reward trajectory groups, it eliminates variance and causes vanishing advantages. This ultimately nullifies policy gradients and discards potentially valuable reasoning trajectories.

Existing research attempts to mitigate these issues via process-level supervision, such as LLM-based reasoning scoring \cite{zhang2025lessons, yuan2025kardia} or search-based strategies like MCTS \cite{zhang2025process, wang2025stepsearch}. Although providing finer-grained feedback, these methods incur heavy overheads from expensive annotations, reliance on proprietary models, and often suffer from limited generalizability across tasks and tools.

To address this dilemma without extra overhead, we propose \textbf{T}urn-level \textbf{S}tage-aware \textbf{P}olicy \textbf{O}ptimization (\textbf{TSPO}). TSPO adopts the First-Occurrence Latent Reward (\textbf{FOLR}) mechanism, which detects the earliest turn containing the ground-truth. For the first facet, TSPO grants partial reward to such turns even if the final answer is wrong, retaining useful process signals. For the second facet, this turn-level allocation increases variance within trajectory groups, avoiding vanishing advantages in uniform-reward groups. Crucially, TSPO achieves this without requiring any external reward models, or additional human labels.

\noindent 
\textbf{Contributions.} In summary, this paper makes the following contributions:
\begin{itemize}[leftmargin=*]
    \item Our analysis and experiments confirm the ``Double Homogenization Dilemma'' (i.e., process and intra-group homogenization) in multi-turn RL and its negative impact.

    \item We propose \textbf{TSPO}, which first locates the earliest occurrence of the ground-truth via the \textbf{FOLR} mechanism, then allocates partial rewards to preserve process signals, and finally performs turn-level advantage estimation,  without requiring any external judges or human annotation overhead.
    \item Extensive experiments across 7 diverse QA datasets demonstrate that TSPO significantly outperforms state-of-the-art baselines, achieving substantial accuracy gains of \textbf{24\%} and \textbf{13.6\%} on Qwen2.5-3B and 7B models, respectively.
\end{itemize}

\section{Double Homogenization Dilemma}
The \textit{Double Homogenization Dilemma} manifests through two coupled facets: process-level and intra-group homogenization. In this section, we analyze each phenomenon and their negative impact.

\subsection{Process-level Reward Homogenization}
\label{sec:process_homo}
Multi-turn QA task typically proceeds in two coupled stages: (1) iterative query reformulation and external information retrieval; and (2) final answer synthesis based on the accumulated evidence. 
Formally, given an input query $x \sim \mathcal{D}$, the agent generates a trajectory:
\begin{equation}
    y = \left[ 
    l_1, f_1, \dots , \underbrace{l_i, f_i}_{i\text{-th turn}},
    \dots,
    l_k, f_k, a
    \right]
\end{equation}
where $l_i$ and $f_i$ represent the reasoning thoughts and environment feedback at turn $i$, and $a$ is the final model answer. 
Let $a_{\mathrm{gold}}$ be the ground-truth answer for $x$. 
Outcome-level RL assigns a sparse outcome reward:
\begin{equation}
    r_{\mathrm{outcome}}(x, y) = \mathbb{I}[a = a_{\mathrm{gold}}] \in \{0, 1\},
\label{eq:traj_reward}
\end{equation}
which \textit{compresses the entire multi-turn reasoning process into a single binary scalar}.

Empirical analysis reveals that $a_{\mathrm{gold}}$ often appears in retrieved feedback $f_i$ at some intermediate turn, yet the model fails to synthesize it correctly, yielding an incorrect final answer (details see Appendix~\ref{sec:case_study}). Under outcome-level supervision, both \textit{retrieval failures} ($a_{\mathrm{gold}} \notin \{f_i\}_{i=1}^k$) and \textit{synthesis failures} ($a_{\mathrm{gold}} \in \{f_i\}_{i=1}^k$ but $a \neq a_{\mathrm{gold}}$) receive identical zero rewards, discarding partial successes in information acquisition.

\begin{figure}[!t]
    \centering
    \includegraphics[width=0.90\linewidth]{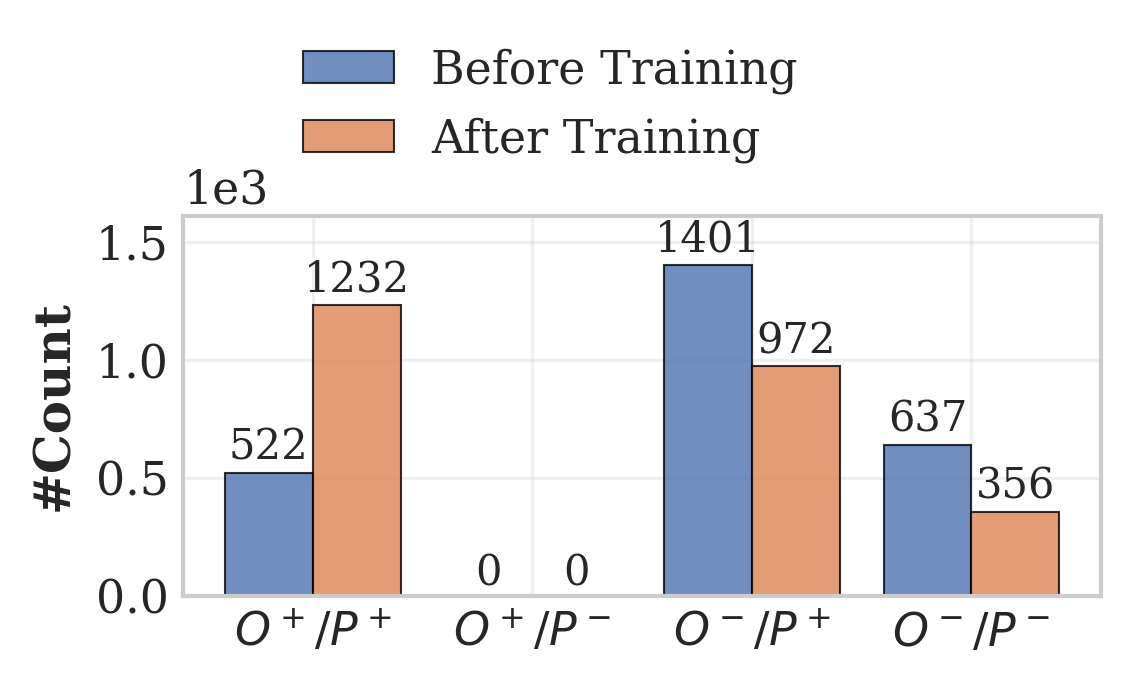}
    \caption{
    Reasoning trajectory distribution ($O$ vs. $P$) for Qwen2.5-7B-Instruct: no retrieval-free successes ($O^{+}/P^{-}$); near-miss ($O^{-}/P^{+}$) and total-failure cases ($O^{-}/P^{-}$ ) treated equally.
    }
    \label{fig:process_homo}
\end{figure}


To quantify this issue, we evaluate trajectories along two orthogonal dimensions:  
\textbf{Outcome Accuracy} ($O$): whether $a = a_{\mathrm{gold}}$; and  
\textbf{Process Integrity} ($P$): whether $a_{\mathrm{gold}}$ appears in any retrieved feedback $\{f_i\}_{i=1}^k$.  
This yields four categories:

\begin{itemize}[leftmargin=*]
    \item \textbf{$O^{-}/P^{-}$}: Complete failure where the model neither retrieves nor synthesizes the correct answer.
    \item \textbf{$O^{-}/P^{+}$}: Near-Miss where correct evidence retrieved but synthesis failed.
    \item \textbf{$O^{+}/P^{+}$}: Full success where both retrieval and synthesis stages are correct.
    \item \textbf{$O^{+}/P^{-}$}: Retrieval-Free Success where the final answer is correct without ever retrieving $a_{\mathrm{gold}}$. 
\end{itemize}

Figure~\ref{fig:process_homo} compares the prevalence of these categories across normal training stages. Two key observations emerge: 
(1) Retrieval-Free Success ($O^{+}/P^{-}$) remains absent, indicating that successful retrieval is a prerequisite for correct synthesis. 
(2) Within failed attempts ($O^-$), both Near-Miss ($O^{-}/P^{+}$) and Total Failure ($O^{-}/P^{-}$) occur but receive the same zero reward. 

This inability to distinguish retrieval success from failure exemplifies \textbf{\textit{process-level reward homogenization}}, where the convergence of heterogeneous intermediate reasoning qualities results in indistinguishable reward signals, erasing fine-grained progress cues and hindering the optimization of tool calling and reasoning.

\subsection{Intra-Group Reward Homogenization}
\label{sec:group_homo}

Group Relative Policy Optimization (GRPO)~\cite{shao2024deepseekmath} stabilizes training by normalizing rewards within groups of trajectories sampled from the same question $x$. For a group of size $G$, let $\{y_i\}_{i=1}^G$ be the sampled trajectories, each assigned a binary outcome reward $r_{\mathrm{outcome},i}(x, y_i) = \mathbb{I}[a_i = a_{\mathrm{gold}}] \in \{0,1\}$. The per-trajectory advantage is computed as:
\begin{equation}
    \hat{A}_i = \frac{r_i - \mu_r}{\sigma_r + \epsilon},
\label{eq:grpo_norm}
\end{equation}
where $\mu_r$ and $\sigma_r$ are the mean and standard deviation of rewards in the group, and $\epsilon$ is a small constant for numerical stability. The policy gradient is proportional to $\hat{A}_i$.



When $\sigma_r = 0$ (i.e., all trajectories in the group receive identical rewards), advantages vanish ($\hat{A}_i \equiv 0$), and the group contributes no gradient updates.
Because \textbf{\textit{process-level reward homogenization}} collapses many Near-Miss ($O^{-}/P^{+}$) trajectories into the same $0$ reward as complete failures, the probability of sampling \textit{uniform-reward groups} (
i.e., $\forall i,j \in \{1,\dots, G\}, \;r_i=r_j \land i\neq j, r_i \in \{1,0\}$
) is greatly increased.

\begin{figure}[!t]
    \centering
    \includegraphics[width=\linewidth]{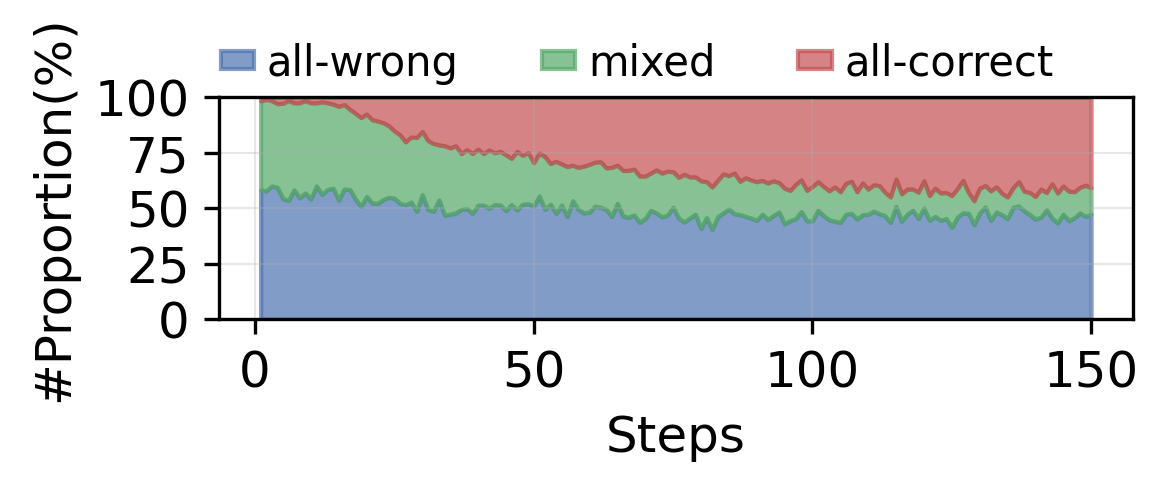}
    \caption{
    Trajectory groups with rollout size $G=5$, composed by \texttt{all-correct}, \texttt{mixed} and \texttt{all-wrong}, during training on Qwen2.5-7B-Instruct. 
    }
    \label{fig:group_homo}
\end{figure}

To quantify this, we empirically quantify the composition of training groups ($G=5$ rollouts) by counting the proportion of all-correct, all-wrong, and mixed groups across training steps' rollouts (Figure~\ref{fig:group_homo}).
Two observations emerge: 
\begin{itemize}[leftmargin=*]
    \item \textbf{High frequency of all-wrong groups}: due to binary rewards, the occurrence of uniform-reward groups is significant, with all-wrong groups maintaining a consistently high proportion ($\geq 40\%$) throughout the normal training. 
    \item \textbf{Stalled optimization from all-wrong groups}: As training progresses, the increase in all-correct groups mainly comes from mixed groups that provide non-zero variance, whereas the proportion of all-wrong groups barely changes.
\end{itemize}

We term this phenomenon \textbf{\textit{intra-group reward homogenization}}: the disappearance of discriminative signals within sampling groups, which prevents the optimizer from exploiting many trajectories and constitutes the second stage of the Double Homogenization Dilemma.


\begin{figure*}
    \includegraphics[width=\linewidth]{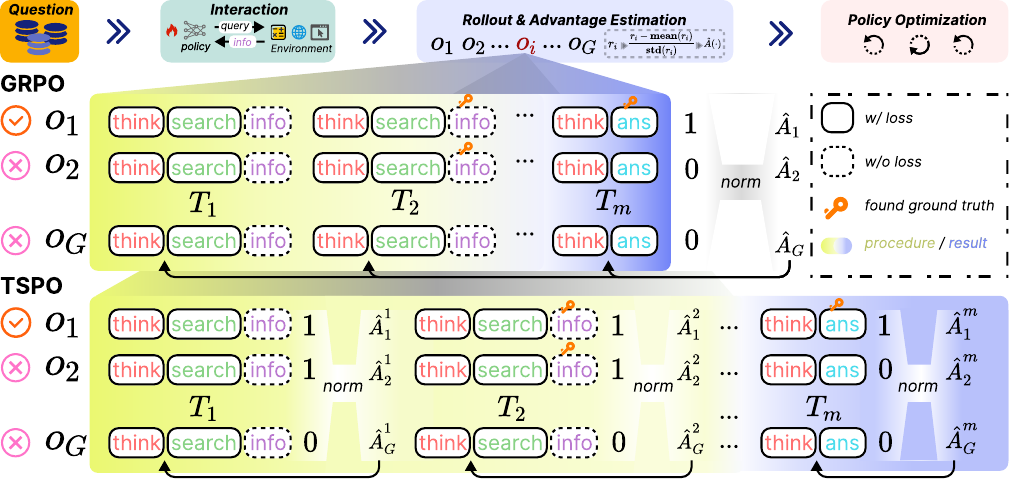}
    \caption{\textbf{Overview of TSPO.} Unlike outcome-level RL which assigns identical zero rewards to $O_2$ ($O^-/P^+$, middle) and $O_G$ ($O^-/P^-$, bottom), TSPO identifies the first occurrence ($t^*$, red key) of $a_{\mathrm{gold}}$ to assign turn-level rewards and performs advantage estimation on a per-turn basis. This restores intra-group variance and provides fine-grained signals. A detailed walkthrough of this example is provided in Section~\ref{par:example}.
    }
    \label{fig:overview}
\end{figure*}

\section{Turn-level Stage-aware Policy Optimization (TSPO)}
To enable fine-grained reward assignment without external supervision, we propose \textbf{Turn-level Stage-aware Policy Optimization (TSPO)}. Unlike prior methods that rely on costly step-level annotations or external reward models~\cite{zhang2025lessons, yuan2025kardia, zhang2025process, wang2025stepsearch}, 
TSPO operates solely on the standard input--answer pairs $(x, a_{\mathrm{gold}})$ and derives turn-level rewards directly from the model's retrieving/reasoning process, thereby addressing the Double Homogenization Dilemma while avoiding additional computation overhead.

\subsection{The First-Occurrence Hypothesis: Identifying Latent Signals}

\paragraph{First-Occurrence Hypothesis}
The presence of $a_{\mathrm{gold}}$ in the intermediate retrieved feedback $f_i$, can serve as a latent signal of partial progress. 
Formally, for any retrieval feedback $f_i \in y$, if $a_{\mathrm{gold}} \in f_i$, then the signal strongly foreshadows the success of the final answer $a$.

\paragraph{Hypothesis Test}
To validate the above hypothesis, we analyze the relationship between \textit{Evidence Presence ($P$)} and \textit{Outcome Accuracy ($O$)} as defined in Section~\ref{sec:process_homo} on 51,713 test trajectories. 
We construct a $2 \times 2$ contingency table (Table~\ref{tab:chi_squared}) that cross-tabulates outcomes along these two binary dimensions.
To test whether $O$ and $P$ are statistically independent, we perform Pearson’s chi-squared test~\cite{plackett1983karl}. 

\begin{table}[h]

\caption{Correlation analysis between outcome accuracy ($O$) and evidence presence ($P$) for Qwen2.5-7B-Instruct }
\label{tab:chi_squared}
\centering
\resizebox{0.45\textwidth}{!}
{%
\begin{tabular}{lccc}
\toprule
 & $O^+$ & $O^-$ & Total Num \\ \midrule
$P^+$ & 10,092 & 25,645 & 35,737 \\
$P^-$ & 0     & 15,976 & 15,976 \\ 
\hline
Total Num & 10,092 & 41,621 & 51,713 \\ \bottomrule
\end{tabular}
}
\end{table}

Under the null hypothesis of independence, the expected frequency for cell $(i,j)$ is:
\begin{equation}
\small
E_{i,j} = 
\frac{(\text{number of $i$-th row}) \times (\text{number of $j$-th column})}{\text{total example number}}
\end{equation}
The chi-squared test statistic is then computed as:
\begin{equation}
\chi^2 = \sum_{i,j\in\{1,2\}} \frac{(O_{i,j} - E_{i,j})^2}{E_{i,j}}
\end{equation}
where $O_{i,j}$ denotes the observed count in cell $(i,j)$ from Table~\ref{tab:chi_squared}.
Substituting the observed and expected frequencies yields $\chi^2 = 5605.5$, which corresponds to a $p$-value less than $0.001$, indicating a highly significant dependency between evidence presence and final correctness.

Notably, all $O^+$ cases (10,092/10,092) also satisfy $P^+$, while $P^-$ cases \emph{never} achieve $O^+$, confirming that retrieval of $a_{\mathrm{gold}}$ is a \emph{necessary condition} for success.
Hence, we incorporate this validated signal, the first-occurrence of the ground-truth into TSPO's reward allocation.

\subsection{The TSPO Framework: Reward Assignment and Optimization}
Building on the First-Occurrence hypothesis, TSPO reformulates sparse outcome rewards into dense, turn-level signals for policy optimization.
\paragraph{Turn-level MDP Formulation.}
We integrate the validated First-Occurrence signal into a turn-level Markov Decision Process (MDP)~\cite{puterman1990markov}, i.e., $\mathcal{M} = \langle \mathcal{S}, \mathcal{A}, \mathcal{T}, \mathcal{R} \rangle$,
where the state $s_k \in \mathcal{S}$ encodes the dialogue history and retrieved evidence up to turn $k$, the action $a_k \in \mathcal{A}$ is either a query to an external tool or a synthesis step, 
$t(s_{k+1} \mid s_k, a_k) \in \mathcal{T}$
is the environment transition updating the reasoning state given $a_k$ and feedback $f_k$, and $r(s_k,a_k) \in \mathcal{R}$ is the turn-level reward defined by the FOLR mechanism.

\paragraph{Reward Assignment with FOLR.}
\label{sec:folr}
Formally, the \textit{First-Occurrence Latent Reward} (FOLR) assigns turn-level reward based on the earliest occurrence of the ground-truth answer in the reasoning trace.

Specifically, given the first-occurrence turn $t^*$ where $a_{\mathrm{gold}}$ appears in trajectory $i$, the reward $r_{i,k}$ for each turn $k$ is defined as:
\begin{equation}
r_{i,k} =
\begin{cases}
1 & \text{if } a_i = a_{\mathrm{gold}}  \\
\alpha & \text{if } a_i \neq a_{\mathrm{gold}} \text{ and } k \le t^*  \\
0 & \text{otherwise}
\end{cases}
\label{eq:folr_formal}
\end{equation}
where $\alpha \in [0,1]$ is a partial reward coefficient. 



\paragraph{Turn-level Advantage Estimation.}
TSPO replaces the trajectory-level scalar $r_i$ in vanilla GRPO with per-turn rewards $r_{i,k}$. The advantage for trajectory $i$ at turn $k$ is computed via group-relative normalization:
\begin{equation}
\hat{A}_{i,k} = \frac{r_{i,k} - \mathrm{mean}(r_{\cdot,k})}{\mathrm{std}({r}_{\cdot,k}) + \epsilon},
\label{eq:advantage}
\end{equation}
where $r_{i,k}$ denotes the set of rewards for all trajectories in the sampling group that contain turn $k$. This turn-level design ensures that Near-Miss ($O_2$) trajectories remain distinguishable from Total Failures ($O_G$), restoring the discriminative gradient signals within the group.

\paragraph{Policy Optimization with TSPO.}
Replacing GRPO's binary $r_i$ and trajectory-level advantage with the FOLR reward in Eq~\eqref{eq:folr_formal} and per-turn advantage Eq~\eqref{eq:advantage}, the optimization objective becomes:


{\small
\begin{equation}
\mathcal{J}_{\mathrm{TSPO}}(\theta) 
= \mathbb{E} \left[ \frac{1}{G} \sum_{i=1}^G \sum_{j=1}^{k} \mathcal{L}_{i,j} 
- \beta D_{\mathrm{KL}}(\pi_\theta \,\|\, \pi_{\mathrm{ref}}) \right].
\label{Eq:optim-obj}
\end{equation}}
where $x\sim \mathcal{D}, y_i \sim \pi_{\theta}$ and the per-turn loss $\mathcal{L}_{i,j}$ is given by

{\small
\begin{equation}
\mathcal{L}_{i,j} 
= \sum \min \left( 
w_{i,t} \hat{A}_{i,j},\; 
\mathrm{clip} \left( w_{i,t},\; 
1\pm\epsilon \right) \hat{A}_{i,j} 
\right).
\label{Eq:ppo-loss}
\end{equation}}
$ w_{i,t} = \frac{\pi_\theta(a_{i,t} \mid s_{i,t})}{\pi_{\text{old}}(a_{i,t} \mid s_{i,t})} $ is the importance coefficient for action $a_{i,t}$ at $t$-th token in turn $k$ of trajectory $i$, and $\hat{A}_{i,k}$ is the normalized per-turn advantage from Eq~\eqref{eq:advantage}.  
By computing advantages and updating policies at the turn level, TSPO preserves informative process signals for \emph{Near-Miss} ($O^-/P^+$) trajectories and introduces reward variance within groups, mitigating both process-level and intra-group reward homogenization.

\paragraph{Example: Turn-level Reward and Advantage}
\label{par:example}
To illustrate how TSPO assigns rewards and computes advantages, consider a sampling group of $G=3$ trajectories for the same question, as depicted in Figure~\ref{fig:overview}. 

As established in Section~\ref{sec:process_homo}, the case O$^+$/P$^-$ (producing the correct answer without retrieving the ground-truth) is empirically absent. 
We therefore concentrate on three observed trajectory types:

\begin{table}[h]
\centering
\caption{Turn-level rewards assigned by TSPO ($\alpha = 1$) for three trajectory types.}
\label{tab:example_rewards}
\resizebox{0.5\textwidth}{!}{
\begin{tabular}{lcccc}
\toprule
Trajectory & Turn 1 & Turn 2 & Turn 3 & Turn 4 \\
\midrule
O$_1$ (O$^+$/P$^+$) & 1 & 1 & 1 & 1 \\
O$_2$ (O$^-$/P$^+$) & 1 & 1 & 0 & 0 \\
O$_G$ (O$^-$/P$^-$) & 0 & 0 & 0 & 0 \\
\bottomrule
\end{tabular}
}
\end{table}

\begin{itemize}[leftmargin=*]
    \item \textbf{Trajectory $O_1$ (O$^+$/P$^+$):} The model retrieves the ground-truth $a_{\text{gold}}$ at turn $t^*$ and correctly synthesizes it in the final answer ($a = a_{\text{gold}}$).
    \item \textbf{Trajectory $O_2$ (O$^-$/P$^+$):} The model retrieves $a_{\text{gold}}$ at turn $t^*$ but fails to produce the correct final answer ($a \neq a_{\text{gold}}$).
    \item \textbf{Trajectory $O_G$ (O$^-$/P$^-$):} The model never retrieves $a_{\text{gold}}$ and produces an incorrect answer.
\end{itemize}

Table~\ref{tab:example_rewards} lists the turn-level rewards assigned by TSPO under partial reward coefficient $\alpha = 1$. Here, ``1'' denotes full or partial reward based on Equation~\eqref{eq:folr_formal}, while ``0'' signifies a zero reward value assigned to that turn.



Turn-level advantages are computed via group normalization. For Turn 1 (rewards = [1, 1, 0]), mean $\mu \approx 0.67$ and $\sigma \approx 0.47$, yielding:
$\hat{A}_{O_{1, 1}} = \hat{A}_{O_{2, 1}} \approx +0.71$, $\hat{A}_{O_{G, 1}} \approx -1.41$.  
For Turn 3 (rewards = [1, 0, 0]), $\hat{A}_{O_{1, 3}} \approx +1.41$, $\hat{A}_{O_{2, 3}} = \hat{A}_{O_{G, 3}} \approx -0.71$.

This shows that TSPO distinguishes Near-Miss ($O_2$) from total failure ($O_G$) by assigning positive rewards up to $t^*$, restoring intra-group variance even in all-wrong groups, while GRPO gives zero advantage to both.



\paragraph{Discussion 1: Variable-length Trajectories.}
For groups with variable numbers of turns, we pad shorter trajectories to the group’s maximum turn using their final-turn reward.  
This allows $\mathrm{mean}(r_{\cdot,k})$ and $\mathrm{std}(r_{\cdot,k})$ to be computed over all $G$ trajectories at each turn, ensuring consistent variance.  
Padded turns are masked during optimization and thus do not affect gradient updates.

\paragraph{Discussion 2: Group Types.}
Groups can be classified as all-correct, all-wrong, and mixed.  
All-correct groups ($O_1$ only) retain trajectory-level allocation as in GRPO.  
All-wrong groups ($O_2$ and $O_G$ only) are the main target of TSPO; per-turn allocation ensures $O_2$ trajectories contribute gradients.  
Mixed groups naturally contain reward variance, and no special handling is applied.  
The focus on all-wrong groups is motivated by their high prevalence, as shown in Section~\ref{sec:group_homo}, and will be further analyzed in our ablation Section~\ref{sec:ablation}.
\begin{table*}[t]
    \centering
    \scriptsize
    \renewcommand{\arraystretch}{1.1}
    \setlength{\tabcolsep}{8pt}
    \caption{Main results. The best performance is set in bold. $^\dagger/^\star$ represents in-domain/out-domain datasets.}\label{tab:main}
    \begin{tabular}{lcccccccc}
        \toprule
        \textbf{Methods} & \multicolumn{3}{c}{\textbf{General QA}} & \multicolumn{4}{c}{\textbf{Multi-Hop QA}} \\
        \cmidrule(r){2-4} \cmidrule(l){5-9}
         & \textbf{NQ$^\dagger$} & \textbf{TriviaQA$^\star$} & \textbf{PopQA$^\star$} & \textbf{HotpotQA$^\dagger$} & \textbf{2wiki$^\star$} & \textbf{Musique$^\star$} & \textbf{Bamboogle$^\star$} & \textbf{Avg.} \\
        \midrule
        \multicolumn{8}{l}{\textbf{Qwen2.5-7b-Instruct}} \\
        Direct Inference & 0.134 & 0.408 & 0.140 & 0.183 & 0.250 & 0.031 & 0.120 & 0.181 \\
        IRCoT & 0.224 & 0.478 & 0.301 & 0.133 & 0.149 & 0.072 & 0.224 & 0.239 \\
        Search-o1 & 0.151 & 0.443 & 0.131 & 0.187 & 0.176 & 0.058 & 0.296 & 0.206 \\
        RAG & 0.349 & 0.585 & 0.392 & 0.299 & 0.235 & 0.058 & 0.208 & 0.304 \\
        SFT & 0.318 & 0.354 & 0.121 & 0.217 & 0.259 & 0.066 & 0.112 & 0.207  \\
        R1-instruct & 0.270 & 0.537 & 0.199 & 0.237 & 0.292 & 0.072 & 0.293 & 0.271  \\
        Rejection Sampling & 0.360 & 0.592 & 0.380 & 0.331 & 0.296 & 0.123 & 0.355 & 0.348 \\
        Search-R1 & 0.393 & 0.610 & 0.397 & 0.377 & 0.404 & 0.146 & 0.368 & 0.385 \\
        ZeroSearch & 0.438 & 0.655 & 0.486 & 0.346 & 0.352 & 0.184 & 0.278 & 0.391 \\
        MT-PPO & 0.498 & 0.649 & 0.459 & 0.428 & 0.404 & 0.215 & - & - \\
        StepSearch & - & - & - & 0.386 & 0.366 & \textbf{0.226} & 0.400 & - \\
        ReasonRAG & - & - & 0.415 & 0.384 & \textbf{0.436} & 0.128 & 0.360 & - \\
        \hdashline
        TSPO & \textbf{0.527} & \textbf{0.687} & \textbf{0.487} & \textbf{0.430} & 0.407 & 0.151 & \textbf{0.416} & \textbf{0.444} \\
        \midrule
        \multicolumn{8}{l}{\textbf{Qwen2.5-3b-Instruct}} \\
        Direct Inference & 0.106 & 0.288 & 0.108 & 0.149 & 0.244 & 0.020 & 0.024 & 0.134 \\
        IRCoT & 0.111 & 0.312 & 0.200 & 0.164 & 0.171 & 0.067 & 0.240 & 0.181 \\
        Search-o1 & 0.238 & 0.472 & 0.262 & 0.221 & 0.218 & 0.054 & 0.320 & 0.255 \\
        RAG & 0.348 & 0.544 & 0.387 & 0.255 & 0.226 & 0.047 & 0.080 & 0.270  \\
        SFT & 0.249 & 0.292 & 0.104 & 0.186 & 0.248 & 0.044 & 0.112 & 0.176  \\
        R1-instruct & 0.210 & 0.449 & 0.171 & 0.208 & 0.275 & 0.060 & 0.192 & 0.224  \\
        Rejection Sampling & 0.294 & 0.488 & 0.332 & 0.240 & 0.233 & 0.059 & 0.210 & 0.265 \\
        Search-R1 & 0.341 & 0.545 & 0.378 & 0.324 & 0.319 & 0.103 & 0.264 & 0.325  \\
        ZeroSearch & 0.414 & 0.574 & 0.448 & 0.274 & 0.300 & 0.098 & 0.111 & 0.317 \\
        StepSearch & - & - & - & 0.351 & 0.328 & \textbf{0.174} & 0.324 & - \\
        ReasonRAG & - & - & 0.329 & 0.300 & 0.266 & 0.069 & 0.136 & - \\
        \hdashline
        TSPO & \textbf{0.490} & \textbf{0.655} & \textbf{0.472} & \textbf{0.384} & \textbf{0.364} & 0.126 & \textbf{0.328} & \textbf{0.403}  \\
        \bottomrule
    \end{tabular}
\end{table*}


\section{Experiments}

\subsection{Experimental Setup}
\paragraph{Datasets}
We evaluate TSPO on seven benchmark datasets, categorized as follows:
(1) \textbf{General Question Answering}: NQ \citep{kwiatkowski2019natural}, TriviaQA \citep{joshi2017triviaqa}, and PopQA \citep{mallen2022not}.
(2) \textbf{Multi-Hop Question Answering}: HotpotQA \citep{yang2018hotpotqa}, 2WikiMultiHopQA \citep{ho2020constructing}, Musique \citep{trivedi2022musique}, and Bamboogle \citep{press2022measuring}.
These datasets provide a diverse range of search with reasoning challenges, enabling a comprehensive evaluation.

\paragraph{Metric}
We report Exact Match score~\cite{maalouly2022exact}, which is the standard metric for QA.

\paragraph{Training Details}
We use Qwen2.5-3B-Instruct and Qwen2.5-7B-Instruct~\cite{team2024qwen2} as the base models, E5~\cite{wang2022text} for retrieval, and the 2018 Wikipedia dump~\cite{karpukhin2020dense} as the corpus.

\paragraph{Baselines}
To evaluate the effectiveness of TSPO, we compare it against the following baselines:
%
(1) \textbf{Inference without Retrieval}: Direct inference and Chain-of-Thought (CoT) reasoning \citep{wei2022chain}.
(2) \textbf{Inference with Retrieval}: Retrieval-Augmented Generation (RAG) \citep{lewis2020retrieval}, IRCoT \citep{trivedi2022interleaving}, and Search-o1 \citep{li2025search}.
(3) \textbf{Fine-Tuning-Based Methods}: Supervised fine-tuning (SFT) \citep{chung2024scaling}, RL-based fine-tuning without a search engine (R1) \citep{guo2025deepseek} and rejection sampling \citep{ahn2024large} with a search engine. 
(4) \textbf{Agentic RL}: Search-R1 \citep{jin2025search}, ZeroSearch \citep{sun2025zerosearch}, MT-PPO~\cite{design2025reinforcing}, 
StepSearch \citep{wang2025stepsearch} and ReasonRAG \citep{zhang2025process}.

\begin{table*}[h!t]
\centering
\caption{\textbf{Ablation study on the Group Type.} Adapt all-wrong groups yields  better performance.}
\resizebox{0.8\textwidth}{!}
{%
\begin{tabular}{ccccccc}

\toprule

{\textbf{Method}} &  {\textbf{Group Type}} & {\textbf{HotpotQA}} & {\textbf{2Wiki}} & {\textbf{MuSiQue}} & {\textbf{Bamboogle}} & {\textbf{Avg.}}\\

 TSPO & \textit{None} & 0.370 & 0.338 & 0.100 & 0.208 & 0.254 \\
 TSPO &\textit{All groups}& \textbf{0.388}&0.363&{0.125}&{0.304}&{0.295}\\
 \rowcolor{lightpurple} TSPO &\textit{All-wrong groups}&{0.384}&\textbf{0.364}&\textbf{0.126}&\textbf{0.328}&\textbf{0.300}\\

\bottomrule
\end{tabular}
}
\label{tab:ablation}
\end{table*}

\subsection{Main Results}
\noindent\textbf{Overall comparison.}
Table~\ref{tab:main} summarizes results on both general QA and multi-hop QA benchmarks.  
Across all datasets, TSPO achieves the highest average performance on both Qwen2.5-7B-Instruct and Qwen2.5-3B-Instruct.  
Compared to the strongest baseline, TSPO delivers relative improvements of \textbf{+13.6\%} and \textbf{+24.0\%} for the 7B and 3B models respectively.  
The consistent gains across in-domain ($\dagger$) and out-of-domain ($\star$) datasets verify the robustness and general applicability of our method.

\paragraph{Compared with sparse outcome-level rewards.}
Relative to outcome-level RL methods such as Search-R1 and ZeroSearch, TSPO attains superior accuracy.  
This benefit stems from alleviating the \textit{Double Homogenization Dilemma}, enabling the model to leverage partial-success trajectories that would otherwise yield zero rewards.  
By converting these trajectories into informative gradient signals, TSPO facilitates more effective policy updates during training.

\paragraph{Compared with dense intermediate rewards.}
Dense-reward methods such as StepSearch, MT-PPO, and ReasonRAG introduce more fine-grained reward assignment than sparse outcome-level RL.  
While these signals improve learning granularity, they typically require more complex infrastructure, curated resources, or heavy computation.  
As shown in Table~\ref{tab:main}, TSPO consistently outperforms all three methods despite using only $(x, a_{\mathrm{gold}})$ supervision.  
This demonstrates that the \emph{First-Occurrence Latent Reward} (FOLR) signal, although simple to obtain, is highly informative and sufficient to drive effective reward assignment.





\section{Further Study}
\subsection{Ablation Study on the Group Types}
\label{sec:ablation}
We conduct an ablation study to examine the impact of applying TSPO's turn-level advantage allocation to different \emph{Group Types} within GRPO.  
Recall from Section~\ref{sec:group_homo} that groups can be classified as \textit{all-correct}, \textit{all-wrong}, or \textit{mixed}. 
We compare three configurations:

\begin{itemize}[leftmargin=*]
    \item \textbf{None}: TSPO reverts to GRPO's trajectory-level advantage allocation, providing no mitigation for process or intra-group reward homogenization.
    \item \textbf{All groups}: turn-level reward assignment is applied to all categories of types.
    \item \textbf{All-wrong groups}: turn-level reward assignment is restricted to all-wrong groups, while all-correct and mixed groups retain trajectory-level rewards.
\end{itemize}

\noindent\textbf{Analysis.}  
The results are presented in Table~\ref{tab:ablation}.
Compared to \textit{None}, both \textit{All groups} and \textit{All-wrong group} achieve substantial gains.  
The drop in \textit{None} is expected: maintaining GRPO's trajectory-level advantage fails to address either process-level or intra-group reward homogenization, leaving partial-success trajectories unused.  

Interestingly, \textit{All-wrong group} slightly outperforms \textit{All groups}. We attribute this to three factors:  
(1) All-correct groups already represent fully successful trajectories and do not benefit from per-turn reward assignment.  
(2) Mixed groups contain intrinsic reward variance, which GRPO can already exploit without modification.  
(3) All-wrong groups are both high in prevalence and homogeneous in reward under GRPO, making them the primary source of wasted samples.  
Thus, selectively enhancing signal in all-wrong groups is sufficient and slightly more effective to unlock the full benefit of FOLR, without unnecessary computation on already-informative groups.


\begin{figure*}
    \centering
    \includegraphics[width=1\linewidth]{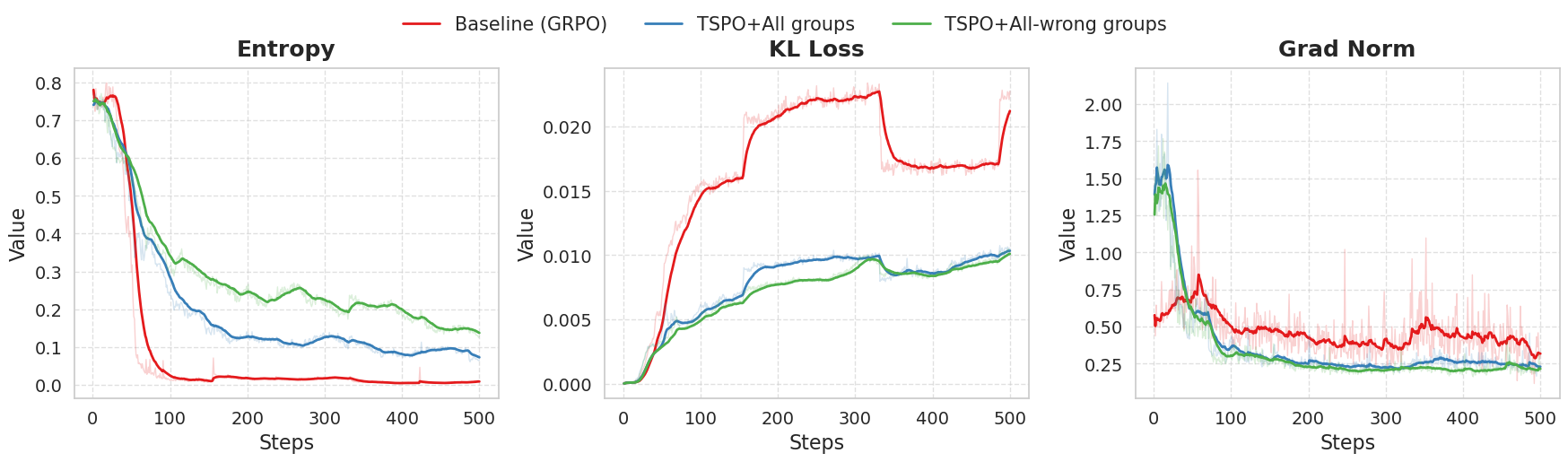}
    \caption{Comparison of Training Dynamics.
    \textbf{Left}: Policy entropy collapses to near zero in GRPO but remains stable under TSPO variants, preserving reasoning diversity. 
    \textbf{Middle}: KL divergence from GRPO spikes in the baseline, indicating unstable policy drift; TSPO maintains consistent alignment. 
    \textbf{Right}: Gradient norms are large and volatile in GRPO due to sparse rewards, while TSPO yields smoother, more consistent updates.}
    \label{fig:dynamics}
\end{figure*}

\subsection{Analysis of Training Dynamics}
\label{sec:dynamics}
To examine how TSPO mitigates the Double Homogenization Dilemma at the optimization level, we track the training dynamics of the policy model across three metrics: policy entropy, KL divergence, and gradient norm. As shown by the logs in Figure~\ref{fig:dynamics}, we compare the baseline (GRPO) with two TSPO variants: \textit{TSPO+All groups} and \textit{TSPO+All-wrong groups}.

\paragraph{Policy Diversity and Entropy Collapse.}
In the baseline, policy entropy decays rapidly to near zero, indicating \textit{policy collapse}: the model overfits to a few trajectories that happen to pass final verification, suppressing exploration of alternative reasoning paths. This behavior stems from the sparsity and binarity of outcome-level rewards. In contrast, both TSPO variants maintain a substantially higher entropy, showing that turn-level stage-aware signals preserve diversity in tool-use and reasoning by attaching value to informative intermediate steps.

\paragraph{Stability and KL Divergence.}
KL divergence measures deviation from the base model’s distribution. The baseline reaches a higher KL value ($\sim$0.02)  with zero entropy, implying drift toward a deterministic yet potentially suboptimal policy. TSPO variants stably converge around 0.01, benefiting from FOLR-based partial rewards that refine search quality without discarding linguistic and reasoning priors. This stabilizes optimization while avoiding sharp shifts away from the base model.

\paragraph{Optimization Consistency via Gradient Norms.}
Gradient norm reflects the magnitude and steadiness of policy updates. The baseline’s higher value ($\sim$0.5) and greater variance align with the intuition that sparse rewards yield vanishing advantages for most samples but cause abrupt updates for the few winners. TSPO reduces and smooths updates ($\sim$0.25) by supplying dense turn-level reward, shaping a steadier optimization landscape. 


\begin{table}[h!t]
\centering
\caption{\textbf{The training time.} We analyze the per-step training time of different methods.}
\resizebox{0.45\textwidth}{!}
{%
\begin{tabular}{ccc}

\toprule

{\textbf{Method}} &  {\textbf{Group Type}} & {\textbf{Time Cost per Step}}  \\

\hline
 \multicolumn{3}{l}{\textbf{\textit{Qwen2.5-3B-Instruct}
}}\\
 Search-R1 & \textit{-} & 243.9s  \\
 TSPO &\textit{All groups}& 278.7s\\
 TSPO &\textit{All-wrong groups}&  268.6s\\

\bottomrule
 \multicolumn{3}{l}{\textbf{\textit{Qwen2.5-7B-Instruct}
}}\\
  Search-R1 & \textit{-} & 414.7s\\
 TSPO &\textit{All groups}&  446.0s\\ 
 TSPO &\textit{All-wrong groups}&  402.3s \\ 

\bottomrule
\end{tabular}
}
\label{tab:time}
\end{table}
\subsection{Training Cost}
\label{app:train_cost}

We compare the per-step training time of TSPO with Search-R1 under two group allocation strategies: \textit{All groups} and \textit{All-wrong groups} (Table~\ref{tab:time}). 
Experiments are conducted with a batch size of 512 on 8 NVIDIA A100 GPUs, with additional training details provided in Appendix~\ref{app:setup}.

Overall, despite introducing turn-level advantage allocation, the training time of TSPO is comparable to that of Search-R1. This is because TSPO does not rely on any external judge or reward model; the proposed FOLR mechanism derives reward allocation directly from $(x, a_{\mathrm{gold}})$ pairs available within the model, without incurring additional inference or annotation cost.
The higher time for \textit{All groups} compared to \textit{All-wrong groups} is due to the increased number of finer-grained turn-level groups that must be processed during advantage calculation. This leads to additional computation in the grouping stage, slightly raising the training cost.

Notably, when using Qwen2.5-7B-Instruct, TSPO on \textit{All-wrong groups} achieves lower per-step time than Search-R1. We attribute this to the fact that Search-R1 assigns identical rewards to all failed trajectories, failing to distinguish their quality. This lack of discriminative feedback hinders learning and results in redundant and repetitive search actions during rollouts, thereby increasing overall training time.

\section{Conclusion}

We identify and formalize the \textit{Double Homogenization Dilemma} in multi-turn tool-augmented reasoning: (1) \textit{process homogenization}, where outcome-only rewards collapse behaviorally distinct trajectories into identical scalar signals; and (2) \textit{intra-group homogenization}, where uniform-reward groups in GRPO yield vanishing gradients. To address both issues, we propose \textbf{TSPO}, a turn-level policy optimization framework centered on the \textbf{First-Occurrence Latent Reward} (FOLR) mechanism. FOLR allocates partial reward to the earliest turn containing the ground truth, requiring only $(x, a_{\mathrm{gold}})$ supervision. Experiments across seven QA benchmarks demonstrate that TSPO consistently outperforms strong baselines, while maintaining computational efficiency.
\section{Limitations}
TSPO shows strong empirical performance and favorable computational properties, but still has limitations:
\begin{itemize}[leftmargin=*]
    \item \textbf{Dependence on retrieval correctness.} FOLR assumes that the correct answer must appear in retrieved evidence. Tasks solvable without direct retrieval may require modified reward definitions.
    \item \textbf{Domain extensibility.} Our experiments focus on \textbf{search-augmented reasoning}. The applicability of TSPO and the Double Homogenization theory to other domains (e.g., code generation, multi-step tool use) remains open.
\end{itemize}
Future work will explore adapting TSPO to broader task types, updating FOLR for settings where retrieval is imperfect or absent.

\bibliography{custom}

\clearpage
\appendix

\section{Related Work}

\paragraph{Tool-integrated Reasoning and Search.}
The integration of external tools, particularly search engines, has significantly enhanced the factuality and multi-hop reasoning capabilities of large language models (LLMs)~\cite{schick2023toolformer, ma2025talk2image, ma2026talk2image}. Early retrieval-augmented generation (RAG) methods focused on single-turn retrieval to alleviate hallucinations~\cite{lewis2020retrieval, xiong2025rag}. Recent advancements have shifted toward agentic and iterative search, where models such as IRCoT~\cite{trivedi2022interleaving} and Search-o1~\cite{li2025search} reformulate queries and reason across multiple turns. ReAct~\cite{yao2022react} demonstrates another effective direction by using prompting to guide iterative reasoning together with search engine calls, enabling models to interleave reasoning steps and tool usage within a single trajectory. While these approaches improve information acquisition, most rely on fixed heuristics or supervised fine-tuning (SFT)~\cite{ouyang2022training}, which may not fully exploit the model's exploratory potential in complex environments.

\paragraph{Reinforcement Learning for Reasoning.}
Reinforcement Learning (RL) has emerged as a powerful paradigm for optimizing reasoning paths~\cite{kaelbling1996reinforcement, ma2025dialogdraw}. Models like DeepSeek-R1~\cite{guo2025deepseek} and Search-R1~\cite{jin2025search} utilize outcome-level rewards to reinforce successful trajectories. Specifically, the Group Relative Policy Optimization (GRPO)~\cite{shao2024deepseekmath} algorithm has gained prominence for its stability and efficiency in group-based normalization. However, relying solely on outcome rewards leads to sparse supervision in long-horizon tasks. Recent work such as DeepResearcher~\cite{zheng2025deepresearcher, dong2025agentic} integrates search agents with real-world web interactions, enriching training signals with dynamic and diverse environments. In addition, other studies~\cite{dong2025tool} explore multi-agent collaboration, where different specialized agents work together to achieve better reasoning and search performance. Our work builds upon the GRPO framework but identifies a critical bottleneck, namely the homogenization of rewards within sampling groups, which hampers the optimization of near-successful attempts.

\paragraph{Process Rewards Assignment.}
To address reward sparsity, researchers have explored process-level supervision~\cite{zhou2025sweet}. Process Reward Models (PRMs)~\cite{zhang2025lessons} and step-wise verifiers provide feedback for individual reasoning steps, although they typically require expensive human annotations or calls to superior teacher models such as GPT-4. Other approaches, such as ReasonRAG~\cite{zhang2025process} and StepSearch~\cite{wang2025stepsearch}, utilize Monte Carlo Tree Search (MCTS)~\cite{browne2012survey} or information-gain metrics to generate dense rewards. Some models~\cite{qian2025toolrl, yang2025qwen3} design explicit process reward mechanisms to evaluate intermediate reasoning steps and tool usage, yet these designs are often tailored to specific tools or datasets, which limits their general applicability. While effective, many of these methods incur significant computational overhead or depend on pre-defined gold evidence~\cite{wang2025spa, feng2025group}. TSPO diverges from these works by extracting a latent "first-occurrence" signal directly from the reasoning trace, enabling fine-grained reward assignment without external judges or additional simulation costs, and providing a lightweight solution for stage-aware policy optimization.

\section{Experimental Setups}
\label{app:setup}
Our implementation is based on \textit{Search-R1}~\cite{jin2025search}, and all training procedures are conducted using the \textit{Verl} framework~\cite{sheng2024hybridflow}. We perform experiments on two model scales: {Qwen-2.5-3B} and {Qwen-2.5-7B}~\cite{team2024qwen2}. 
We train each model for a total of 500 steps. For GRPO training, the group size ($G$) is set to $5$. The entropy coefficient is fixed at $0$, and the KL-loss coefficient is set to $1\times10^{-3}$. 
Training is performed on {8 NVIDIA A100 GPUs}. The total batch size is $512$, with a mini-batch size of $256$ and a micro-batch size of $8$ per GPU.


\section{Resource Details}
Table~\ref{tab:dataset_details} summarizes all resources used in our experiments, including datasets, models, and reference codebases. 
For each dataset, we list the number of examples and provide the corresponding URL for public access. 
The seven QA datasets cover both single-hop and multi-hop scenarios, as well as in-domain and out-of-domain settings, enabling a diverse evaluation of search-augmented reasoning. 
The model entries include the Qwen2.5-3B-Instruct and Qwen2.5-7B-Instruct variants, which serve as our base policy models during reinforcement learning. 
Reference libraries contain open-source implementations used in training, including Verl for RLHF infrastructure and Search-R1 as the baseline framework for our experiments.
All listed resources are publicly available, ensuring that our setup can be reproduced without reliance on proprietary data or closed-source software.

\begin{table*}[h]
\small
\centering
\caption{Overview of datasets, models, and reference libraries used in our experiments, including sizes and URLs.}
\label{tab:dataset_details}
\renewcommand{\arraystretch}{1.1}
\scalebox{0.9}{
\setlength{\tabcolsep}{1.55mm}
\begin{tabularx}{\linewidth}{l l X}
\toprule
\textbf{Type} & \textbf{Size/Params} & \textbf{URL} \\
\midrule

\cellcolor[HTML]{EFF0F0} \gray{\textit{Dataset}}&
\cellcolor[HTML]{EFF0F0} &
\cellcolor[HTML]{EFF0F0} \\


Natural Questions (NQ)~\cite{kwiatkowski2019natural} & 100,231 &
\tiny{\url{https://ai.google.com/research/NaturalQuestions}} \\

TriviaQA~\cite{joshi2017triviaqa} & 847,579 
& \tiny{\url{https://huggingface.co/datasets/mandarjoshi/trivia_qa}} \\

PopQA~\cite{mallen2022not} & 14,267 
& \tiny{\url{https://huggingface.co/datasets/akariasai/PopQA}} \\

HotpotQA~\cite{yang2018hotpotqa} & 203,109 
& \tiny{\url{https://hotpotqa.github.io/}} \\

2WikiMultiHopQA~\cite{ho2020constructing} & 192,606
& \tiny{\url{https://github.com/Alab-NII/2wikimultihop}} \\

Musique~\cite{trivedi2022musique} & 22,355
& \tiny{\url{https://github.com/stonybrooknlp/musique}} \\

Bamboogle~\cite{press2022measuring} & 125 
& \tiny{\url{https://huggingface.co/datasets/cmriat/bamboogle}} \\

\midrule
\cellcolor[HTML]{EFF0F0} \gray{\textit{Model}}&
\cellcolor[HTML]{EFF0F0} &
\cellcolor[HTML]{EFF0F0} \\

Qwen2.5-3B-Instruct~\cite{team2024qwen2} & 3B &
\tiny{\url{https://huggingface.co/Qwen/Qwen2.5-3B-Instruct}} \\

Qwen2.5-7B-Instruct~\cite{team2024qwen2} & 7B &
\tiny{\url{https://huggingface.co/Qwen/Qwen2.5-7B-Instruct}} \\

\midrule
\cellcolor[HTML]{EFF0F0} \gray{\textit{Reference library}}&
\cellcolor[HTML]{EFF0F0} &
\cellcolor[HTML]{EFF0F0} \\

Verl~\cite{sheng2024hybridflow} & - & 
\tiny{\url{https://github.com/volcengine/verl}} \\

Search-R1~\cite{jin2025search} & - & 
\tiny{\url{https://github.com/PeterGriffinJin/Search-R1}} \\

\bottomrule
\end{tabularx}
}
\end{table*}

\section{Train Details}

\subsection{Training Prompt}
\begin{promptbox}[label=box: training prompt]{Training Prompt}
Answer the given question. \
You must conduct reasoning inside \think{and} first every time you get new information. \
After reasoning, if you find you lack some knowledge, you can call a search engine by \call{\{"name": <function-name>, "arguments": <args-json-object>\}} and it will return the top searched results between \response{and}. You can search as many times as your want. \
If you find no further external knowledge needed, you can directly provide the answer inside \answer{and}, without detailed illustrations. For example, \answer{Beijing}. Question:
\end{promptbox}

\begin{figure*}
    \centering
    \includegraphics[width=1\linewidth]{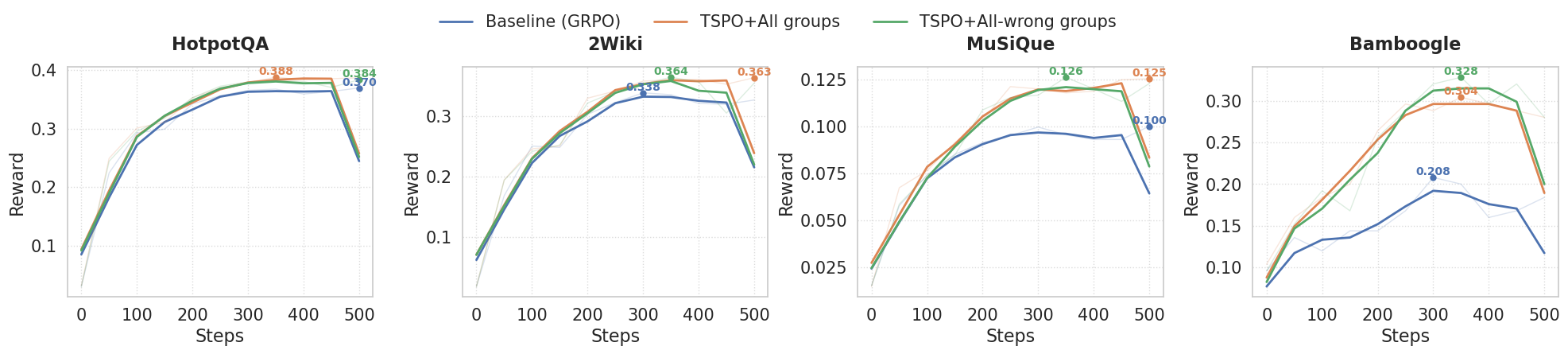}
    \caption{Training reward curves on four representative datasets. TSPO variants show consistently faster and more stable reward convergence compared to the GRPO baseline.}
    \label{fig:train_rewards}
\end{figure*}
\subsection{Train Rewards}
\label{app:train_rewards}
Figure~\ref{fig:train_rewards} shows the training reward curves of three methods: GRPO (outcome-level baseline), TSPO with turn-level rewards applied only to \textit{all-wrong groups}, and TSPO applied to \textit{all groups} across four multi-hop QA datasets: HotpotQA, 2WikiMultiHopQA, MuSiQue, and Bamboogle.

We observe that GRPO exhibits slow and unstable reward growth across all datasets, often plateauing early. This reflects the Double Homogenization Dilemma: sparse binary rewards cause most trajectory groups (especially all-wrong ones) to yield zero advantages, severely limiting learning signal. In contrast, both TSPO variants achieve consistently faster and smoother reward convergence. Notably, even though TSPO+All-wrong modifies rewards only in homogeneous failure groups, it matches or nearly matches the performance of TSPO+All groups, which confirmes our analysis in Section~\ref{sec:ablation} that all-wrong groups are the primary bottleneck in standard GRPO training.

The accelerated reward learning under TSPO demonstrates that leveraging first-occurrence latent signals effectively transforms previously wasted near-miss trajectories into useful supervision, enabling more sample-efficient policy optimization.

\subsection{Reasoning Efficiency and Response Length}
\label{sec:efficiency}
We monitor the average response length (in tokens) throughout training. All models begin with verbose reasoning trajectories (over 2,100 tokens) but gradually converge to more concise outputs (1,700–1,900 tokens). Crucially, while outcome-based RL baselines often resort to aggressive shortening that sacrifices necessary retrieval or reasoning steps, TSPO variants achieve compression through improved reasoning efficiency. This indicates that turn-level supervision guides the policy to eliminate redundancy without discarding critical intermediate actions.

\begin{figure}[!t]
    \centering
    \includegraphics[width=\linewidth]{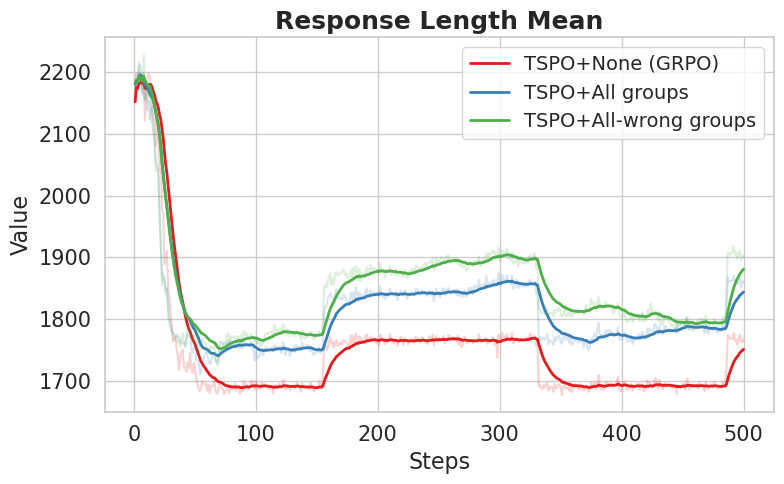}
    \caption{\textbf{Response length evolution during training.} TSPO variants reduce trajectory length steadily while preserving reasoning integrity, avoiding the premature truncation observed in outcome-only methods.}
    \label{fig:response_length}
\end{figure}

\subsection{Training Algorithm of TSPO}
\label{sec:algo}

We present the complete training procedure of \textit{Turn-level Stage-aware Policy Optimization} (TSPO), which integrates the First-Occurrence Latent Reward (FOLR) mechanism into a group-based reinforcement learning framework. TSPO supports two practical advantage allocation strategies: \texttt{All groups}, which applies turn-level normalization to all trajectory groups, and \texttt{All-wrong groups}, which applies it only to homogeneous failure groups. This selective design is motivated by our analysis in Section~\ref{sec:group_homo}, which shows that all-wrong groups are the primary source of wasted samples under standard GRPO.

As outlined in Algorithm~\ref{app:algo}, TSPO proceeds in three stages per training batch. First, it performs multi-turn rollouts to collect diverse reasoning trajectories for each question. Second, it assigns turn-level rewards using FOLR: unsuccessful trajectories receive partial reward ($\alpha$) for all turns up to and including the first appearance of $a_{\text{gold}}$, while successful trajectories retain full outcome reward across all turns. Third, it computes advantages either at the turn level or the trajectory level, depending on the group composition and the chosen strategy. This enables TSPO to preserve fine-grained signals from near-miss attempts and mitigate both process homogenization and intra-group homogenization.

\begin{algorithm}[h]
\caption{TSPO Training Stage}
\normalsize
\begin{algorithmic}[1]
\REQUIRE Policy $\pi_\theta$, Reference $\pi_{\text{ref}}$, Dataset $\mathcal{D}$, Ground-truth $a_{\text{gold}}$, Group size $G$, Reward coeff. $\alpha$, Strategy $\in \{\texttt{All-groups}, \texttt{All-wrong}\}$
\ENSURE Optimized policy $\pi_\theta$

\STATE \textbf{Rollout:} Sample batch $\mathcal{D}_b \subset \mathcal{D}$
\FOR{$q \in \mathcal{D}_b$}
    \STATE Sample $G$ trajectories $\{y_i\}_{i=1}^G \sim \pi_\theta(q)$
    \STATE \textbf{FOLR:} 
    \FOR{$y_i$}
        \STATE $r^{\text{out}}_i \gets \mathbb{I}(\text{EM}(y_i, a_{\text{gold}}))$
        \STATE $t^* \gets \text{first turn with } a_{\text{gold}} \in \text{Response}_{i,t}$
        \FOR{$t=1$ to $T_i$}
            \STATE $r_{i,t} \gets \begin{cases} \alpha & \text{if } t \le t^* \land r^{\text{out}}_i=0 \\ r^{\text{out}}_i & \text{otherwise} \end{cases}$
        \ENDFOR
    \ENDFOR
    \STATE \textbf{Advantage Estimation:}
    \STATE $\bar{r} \gets \frac{1}{G} \sum r^{\text{out}}_i$
    \IF{Strategy = \texttt{All-wrong} \AND $\bar{r}=0$}
        \FOR{$t$}
            \STATE $A_{i,t} \gets \frac{r_{i,t} - \text{mean}(\{r_{j,t}\})}{\text{std}(\{r_{j,t}\})+\epsilon}$
        \ENDFOR
    \ELSIF{Strategy = \texttt{All-groups}}
        \STATE Compute turn-level $A_{i,t}$ for all groups
    \ELSE
        \STATE $A_i \gets \frac{r^{\text{out}}_i - \text{mean}(r^{\text{out}})}{\text{std}(r^{\text{out}})+\epsilon}$
    \ENDIF
\ENDFOR

\STATE \textbf{Policy Update:} Optimize $\pi_\theta$ w.r.t. advantages $\{A_{i,t}\}$
\RETURN $\pi_\theta$
\end{algorithmic}
\label{app:algo}
\end{algorithm}

\section{Hyperparameter Analysis}
\label{subsec:hyperparam}
\begin{figure}[!t]
  \centering
  \includegraphics[width=1\linewidth]{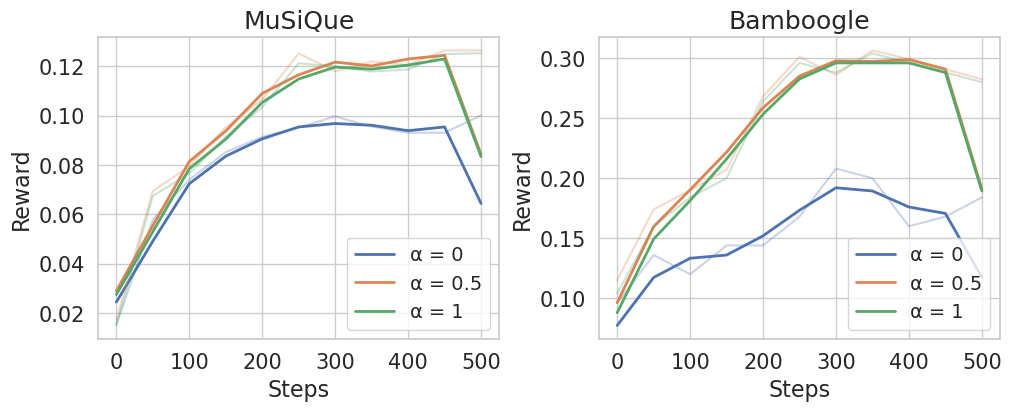} 
  \caption{\textbf{Hyperparameter analysis of $\alpha$ in FOLR.} Results show that any $\alpha > 0$ effectively mitigates intra-group homogenization, with negligible difference between $\alpha = 0.5$ and $\alpha = 1$.}
  \label{fig:hyper}
\end{figure}

We evaluate the influence of the partial reward coefficient $\alpha$ in the FOLR mechanism (Section~\ref{sec:folr}), considering $\alpha \in \{0, 0.5, 1\}$. The results are presented in Figure~\ref{fig:hyper}.

When $\alpha = 0$, TSPO degenerates into standard GRPO since all incorrect trajectories are assigned zero rewards across all turns. This leads to intra-group homogenization in all-wrong groups, where Near-Miss trajectories ($O_2$) and complete failures ($O_G$) become indistinguishable in terms of reward signals. Such homogenization erases potential process-level cues, causes vanishing advantages, and ultimately yields the lowest performance.

For $\alpha = 0.5$ and $\alpha = 1$, intra-group variance is restored by assigning positive rewards to turns preceding the first occurrence of $a_{\mathrm{gold}}$ in Near-Miss ($O_2$) trajectories. The average performances are similar because group-wise normalization (Equation~\ref{eq:advantage}) preserves the relative ordering of trajectories, where $O_2$ consistently obtains a normalized advantage around $+1$ whereas $O_G$ obtains approximately $-1$. This indicates that any $\alpha > 0$ is sufficient to mitigate intra-group reward homogenization, with the precise value having minimal effect on final performance.

\section{Case Study: Partial Successes Penalized as Total Failures}
\label{sec:case_study}
Under outcome-level supervision, any deviation from the exact ground-truth answer results in a zero reward, regardless of whether critical evidence was successfully retrieved. This collapses trajectories with meaningful partial progress into the same failure category as those with no progress at all.

We present two such failure cases from our training logs (Table~\ref{tab:case_interaction}). In the first, the model is asked for the tissue type of the innermost layer of cells. The correct answer is \textit{Epithelium}. After one search turn, multiple documents (Docs 1, 2, 4) explicitly state that ``Epithelium is one of the four basic types of animal tissue'', directly providing $a_{\mathrm{gold}}$. However, the model misinterprets biological context and outputs \textit{Endoderm}, confusing germ layers with tissue types.

In the second case, the model searches for the first Nobel Prize in Physics winner. The retrieved results include both ``Wilhelm Röntgen'' (Doc 1) and the full name ``Wilhelm Conrad Röntgen'' (Doc 3). Despite having access to the complete answer, the model outputs only ``Wilhelm Röntgen'', which fails under Exact Match evaluation.

Critically, both trajectories successfully acquire the ground-truth information during retrieval, yet are penalized identically to trajectories that never retrieve relevant evidence. This exemplifies \textbf{process-level reward homogenization}: diverse intermediate successes are erased by a single binary outcome signal. TSPO addresses this by assigning partial reward to the turn where $a_{\mathrm{gold}}$ first appears, preserving signals of effective search behavior even when final synthesis is imperfect.

\begin{table*}[t]
\centering
\caption{Two failure trajectories where the ground-truth answer appears in retrieved documents (highlighted in red), but the final output is incorrect due to reasoning errors (top) or incomplete answer formatting (bottom). Both receive zero reward under outcome-level supervision despite successful evidence retrieval.}
\footnotesize
\begin{tabular}{p{13.5cm}}





\toprule

\textbf{Question}: name the tissue type of the inner most layer of cells? \\
\textbf{Answer}: Epithelium \\
\textbf{Qwen2.5-7b-ins}: \\
\think{he question is asking about the tissue type of the innermost layer of cells. This typically refers to the lining of an organ or a body cavity. However, without more context, it's difficult to determine which specific organ or body part is being referred to. I will conduct a search to find the most common innermost tissue type.}

\call{\{"name": "search", "arguments": \{"query\_list": ["innermost layer of cells tissue type"]\}\}}

\response{Doc 1 \textcolor{red}{(Title: Epithelium)Epithelium Epithelium () is one of the four basic types of animal tissue, along with connective tissue, muscle tissue and nervous tissue.} Epithelial tissues line the outer surfaces of organs and blood vessels throughout the body, as well as the inner surfaces of cavities in many internal organs... 
Doc 2: \textcolor{red}{(Title: Epithelium)form is epithelia...} There are three principal shapes of epithelial cell: squamous, columnar, and cuboidal. These can be arranged in a single layer of cells as simple epithelium, either squamous, columnar, or cuboidal, or in layers of two or more cells..\\
Doc 3: (Title: Periosteum)Periosteum The periosteum is a membrane that covers the outer surface of all bones, except at the joints of long bones...
Doc 4: \textcolor{red}{(Title: Anatomy)connective tissue in the extracellular matrix secreted by the epithelial cells.} There are many different types of epithelium, modified to suit a particular function...
Doc 5: \textcolor{pink}{(Title: Endoderm)Endoderm Endoderm is one of the three primary germ layers in the very early embryo.} The other two layers are the ectoderm (outside layer) and mesoderm (middle layer), with the endoderm being the innermost layer...
}

\think{Based on the search results, the innermost layer of cells is typically referred to as the endoderm. The endoderm is one of the three primary germ layers in the very early embryo, along with the ectoderm and mesoderm. It forms the epithelial lining of multiple systems. Therefore, the tissue type of the innermost layer of cells is:<answer> Endoderm </answer>} \\
\answer{Endoderm} \\

\midrule
\textbf{Question}: who got the first nobel prize in physics? \\
\textbf{Answer}: Wilhelm Conrad Röntgen \\
\textbf{Qwen2.5-7b-ins}: \\
\think{Let me search for the information about the first Nobel Prize in Physics winner.}

\call{\{"name": "search", "arguments": \{"query\_list": ["first Nobel Prize in Physics winner"]\}\}}

\response{Doc 1 (Title: "Nobel Prize in Physics")The Nobel Prize in Physics () is a yearly award given by the Royal Swedish Academy of Sciences for those who have made the most outstanding contributions for mankind in the field of physics>... \textcolor{pink}{The first Nobel Prize in Physics was awarded to physicist Wilhelm Röntgen in recognition of the extraordinary services...} 
Doc 2: (Title: "Nobel Prize in Physics")receive a diploma, a medal and a document confirming the prize amount...
Doc 3: (Title: "Wilhelm Röntgen")...\textcolor{red}{Wilhelm Conrad Röntgen (; ; 27 March 1845 – 10 February 1923) was a German mechanical engineer and physicist ... earned him the first Nobel Prize in Physics in 1901...}
}

\think{Based on the search results, the first Nobel Prize in Physics was awarded to Wilhelm Röntgen in recognition of the extraordinary services he rendered by his discovery of X-rays. Therefore, the answer is <answer>Wilhelm Röntgen</answer>.} \\
\answer{Wilhelm Röntgen} \\
\bottomrule
\end{tabular}

\label{tab:case_interaction}
\end{table*}

\end{document}